\documentclass[runningheads]{llncs}

\usepackage{eccv}

\usepackage{eccvabbrv}
 \usepackage {diagbox}
\usepackage{subcaption}
\usepackage{graphicx}
\usepackage{booktabs}
\usepackage{comment}
\usepackage[accsupp]{axessibility}  

\usepackage{amsmath}
\usepackage{amssymb}
\usepackage{times}
\usepackage{epsfig}
\usepackage{arydshln}
\usepackage{multirow}
\usepackage{subcaption}
\usepackage{musicography}
\usepackage[flushleft]{threeparttable}
\usepackage{capt-of,etoolbox}
\usepackage{subcaption}
\usepackage{wrapfig}

\usepackage[pagebackref,breaklinks,colorlinks,citecolor=eccvblue]{hyperref}

\usepackage{orcidlink}
  
\newcommand{\yy}[1]{{\color{red} #1}}

\begin{document}

\title{DiffSurf: A Transformer-based Diffusion Model for \\  Generating and Reconstructing 3D Surfaces in Pose} 

\titlerunning{DiffSurf}

\author{Yusuke Yoshiyasu\inst{1}\orcidlink{0000-0002-0433-9832} \and
Leyuan Sun\inst{1}\orcidlink{0000-0001-6123-9339}}

\authorrunning{Y.Yoshiyasu and L. Sun}

\institute{National Institute of Advanced Industrial Science and Technology (AIST), 1-1-1 Umezono, Tsukuba, Japan 
\email{\{yusuke-yoshiyasu,son.leyuansun\}@aist.go.jp}}

\maketitle

\begin{abstract}
This paper presents DiffSurf, a transformer-based denoising diffusion model for generating and reconstructing 3D surfaces. Specifically, we design a diffusion transformer architecture that predicts noise from noisy 3D surface vertices and normals. With this architecture, DiffSurf is able to generate 3D surfaces in various poses and shapes, such as human bodies, hands, animals and man-made objects. Further, DiffSurf is versatile in that it can address various 3D downstream  tasks including morphing, body shape variation and 3D human mesh fitting to 2D keypoints. Experimental results on 3D human model benchmarks demonstrate that DiffSurf can generate shapes with greater diversity and higher quality than previous generative models. Furthermore, when applied to the task of single-image 3D human mesh recovery, DiffSurf achieves accuracy comparable to prior techniques at a near real-time rate. \url{https://github.com/yusukey03012/DiffSurf}
\end{abstract}

\begin{figure}[t]
\begin{center}
 \includegraphics[width=1\linewidth]{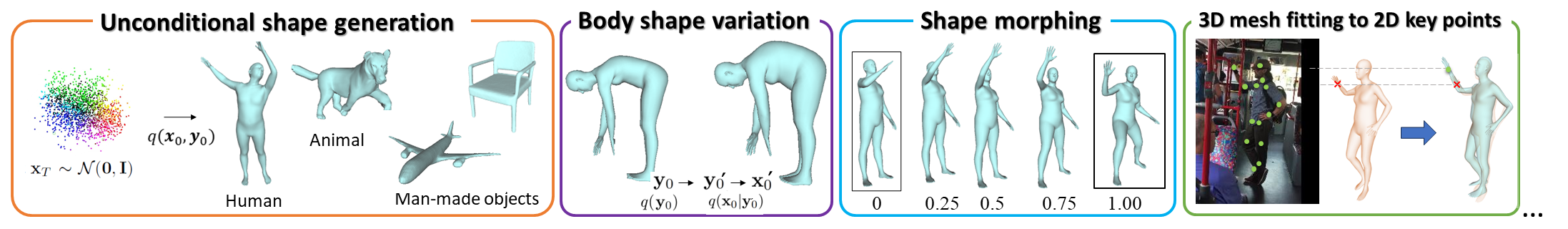}
  \vspace{-20pt}
 \caption{DiffSurf addresses the unconditional generation of 3D surfaces in diverse poses. It can generate 3D surfaces of various objects types such as humans, mammals and man-made objects. Downstream tasks, including unconditional generation, morphing and fitting to 2D key points can be addressed with pre-trained DiffSurf models. }
\vspace{-20pt}
 \label{fig:teaser}
\end{center}
\end{figure}

\section{Introduction}
\label{sec:intro}

Creating and reconstructing 3D shape models in various shapes and poses is a significant challenge in computer vision and computer graphics, with extensive applications in gaming, augmented reality (AR) and  virtual reality (VR). For such 3D content-based applications, a surface mesh is the most commonly used shape representation because of its efficiency during graphics rendering and user-friendliness for artists.


Over the past few years, diffusion models \cite{ho2020denoising, Rombach_2022_CVPR} have revolutionized the content creation paradigm in the image domain, particularly in the task of image generation from text prompts. Diffusion models can generate high-quality and diverse data by learning to reverse the diffusion process. This process gradually constructs desired data samples from noise whose dimensionality is higher than that of previous generative models such as generative adversarial networks (GANs). Additionally, diffusion models have been applied to the generation of 3D data, such as object point clouds \cite{luo2021diffusion, zeng2022lion,slide2023},  textured 3D models  \cite{poole2022dreamfusion,lin2023magic3d, wang2023prolificdreamer}, scene radiance field \cite{bautista2022gaudi} and 3D human pose \cite{gong2023diffpose}. Recent 3D shape diffusion models are able to generate 3D surfaces with complex geometry and topology by incorporating implicit functions, such as signed distance fields (SDF), and extracting their zero level-set surface using the marching cubes algorithm \cite{shim2023diffusion,cheng2023sdfusion}.

Yet, there remain several challenges in generating 3D surfaces based on diffusion models. {\bf Firstly}, the current approaches do not consider pose. In the case of 3D human and animal generation in different poses, point-to-point correspondences between shapes are important but they are lost when employing the current 3D shape diffusion models. There exist recent works of diffusion models for generating 3D human poses and shapes conditioned on images  \cite{dat2023,cho2023generative,li2023diffhand,gong2023diffpose} but  generation of diverse body shapes and poses are not considered. {\bf Secondly}, we want our generative model to handle a wide range of objects, such as human bodies, mammals and man-made objects.  {\bf Thirdly}, a framework that can deal with a wide variety tasks, e.g. interpolating two shapes, altering pose and manipulating shape, is highly sought after.

In this paper, we propose a transformer-based diffusion model for generating and reconstructing 3D surfaces (dubbed DiffSurf). To address the aforementioned challenges,  we design a diffusion transformer architecture that predicts noise from noisy 3D surface vertex coordinates. By representing a surface with points and normals, processing them in diffusion transformer and then employing up-samplers dedicated for topologically fixed and varied cases, DiffSurf is able to handle diverse body poses, various object types and multiple different tasks as illustrated in Figs. \ref{fig:teaser} and \ref{fig:overview}. To our knowledge, DiffSurf is the first diffusion model that addresses generation of 3D surfaces in diverse body poses and shapes. The contributions of this paper are summarized as follows:


\begin{enumerate}
\item DiffSurf, a denoising diffusion transformer model that can generate 3D surfaces in various body shapes and poses.

\item It can generate 3D shapes of diverse object types, such as human bodies, mammals and man-made objects, using a diffusion transformer model that leverages point-normal representation. 

\item It provides methodologies for addressing various 3D processing and image-to-3D downstream tasks by effectively utilizing pre-trained DiffSurf models based on score distillation sampling (SDS) and classifier-free guidance (CFG). 

\end{enumerate}

\section{Related Work}
\label{sec:related}
\noindent {\bf Generative models for 3D shape and pose } 
Previous generative models for 3D shape and pose generation predominantly utilize variational autoencoders (VAEs) \cite{8578710, yuan2020mesh, hierachical2020, LIMP2020, Jiang2020HumanBody, 9010824,zhou20unsupervised, SemanticHuman}, generative adversarial networks (GANs) \cite{hmrKanazawa17, 9879900} or normalizing flows  \cite{xu2020ghum, biggs2020multibodies, zanfir2020weakly}. SMPL-X \cite{SMPL-X:2019} employs a VAE to learn a pose prior and enforces constraints on body joint angles. COMA \cite{COMA:ECCV18} and CAPE \cite{ma2020cape} designed their VAEs based on graph neural networks to model facial expressions and clothing geometric deformations, respectively. 
Kanazawa et al. \cite{hmrKanazawa17} introduced the human mesh recovery (HMR) technique that estimates a posed human body model from a single image by employing GANs to to provide body pose and shape priors. Other approaches focus on designing and obtaining latent encoders or representation using GANs \cite{9879900,DBLP:journals/corr/abs-1903-10384,iepgan2021}. Recent techniques for 3D human pose and shape estimation employ normalizing flows \cite{xu2020ghum, biggs2020multibodies, zanfir2020weakly,kolotouros2021prohmr} in attempting to learn 3D priors from motion capture datasets. 
Pose-NDF instead represents and learns a pose space using neural distance fields \cite{tiwari22posendf}.

\noindent {\bf Diffusion models and 3D generation } In 3D generation, Luo et al. \cite{luo2021diffusion} proposed the first 3D point cloud generation method based on diffusion models by extending Denoising Diffusion Probabilistic Models (DDPM) \cite{ho2020denoising}. LION \cite{zeng2022lion} and SLIDE \cite{slide2023} adopted Latent Diffusion Models (LDM) \cite{Rombach_2022_CVPR} in point cloud generation, aiming to reduce point cloud resolution for more efficient training and sampling. A learning-based surface reconstruction technique called ShapeAsPoint (SAP) \cite{Peng2021SAP} is then employed to convert the generated point clouds into volume functions and subsequently into meshes. To generate 3D surfaces, recent approaches use implicit functions in diffusion process such as SDF, and extract a mesh from 3D volume using marching cubes \cite{cheng2023sdfusion, yu2023surf, shim2023diffusion}. MeshDiffusion \cite{Liu2023MeshDiffusion} uses DMTet \cite{shen2021dmtet} that combines a tetra mesh and SDF to represent the object shape to generate topologically and geometrically complex objects.  Mo et al. proposed DiT-3D \cite{mo2023dit3d} which extends diffusion transformer \cite{Peebles2022DiT, Peebles2022, bao2022one, bao2022all} to  voxelized 3D point clouds, accomplishing the generation of 3D objects. PolyDiff introduced a 3D diffusion model that can work with polygonal meshes \cite{alliegro2023polydiff}. Point-e \cite{nichol2022pointe} and Shape-e \cite{jun2023shape} extend \cite{luo2021diffusion} to colored point clouds using transformers. 

\noindent {\bf 3D shape and pose from image } Human mesh recovery approaches \cite{tian2022hmrsurvey} predict a 3D human body mesh from a single image or video frames, which can be roughly divided into parametric \cite{bogo2016keep, hmrKanazawa17, pymaf2021} and vertex-based approaches \cite{kolotouros2019cmr, Choi_2020_ECCV_Pose2Mesh, lin2021end-to-end}. Parametric approaches regress the body shape and pose parameters of human body models like SMPL or SMPL-X. On the other hand, vertex-based approaches directly regress from an image to 3D vertex coordinates. Transformer architectures, which are known for their ability to capture long-range dependencies, have been employed in vertex-based human mesh recovery and shown strong performances \cite{lin2021end-to-end, lin2021-mesh-graphormer, cho_arxiv.2207.13820,10096870,yoshiyasu2023-deformer}. The reconstruction of mammals from images has also been addressed in the field \cite{xu2023animal3d, Zuffi:CVPR:2017, BARC:2022}. Research on 3D human body pose and shape generation via diffusion models has recently commenced but most being task-specific and conditional on 2D data. They include 3D human pose estimation methods from 2D keypoints \cite{gong2023diffpose, shan2023diffusion}, parametric human mesh recovery techniques  \cite{10161247, cho2023generative} and vertex-based human mesh recovery methods  \cite{li2023diffhand, dat2023}.

\section{Background:  Diffusion models}

Diffusion models establish a Markov chain of diffusion steps by gradually adding random noise to data (i.e., a forward diffusion process) and learn to reverse this process to construct desired data samples from the noise (i.e., a reverse diffusion process). Considering a data point sampled from a data distribution ${\bf x}_0 \sim q({\bf x})$, the forward process produces a sequence of noisy samples ${\bf x}_1\ldots{\bf x}_T$ by adding a small amount of Gaussian noise to the sample in $T$ steps:
\begin{align}
q({\bf x}_{1:T} | {\bf x}_0) &= \prod_{t=1}^T q({\bf x}_{t} | {\bf x}_{t-1}) \\
q({\bf x}_{t} | {\bf x}_{t-1}) &= \mathcal{N} ({\bf x}_{t} ; \sqrt{1 - \beta_t} {\bf x}_{t-1}, \beta_t {\bf I})    
\end{align}
where $\{\beta_t \}_1^T \in (0,1)$ is the noise variance schedule. 

Data generation can be initiated from a Gaussian noise input ${\bf x}_T \sim \mathcal{N}({\bf 0},{\bf I})$, provided that the aforementioned forward process is reversed to sample from $q({\bf x}_{t-1} | {\bf x}_{t})$. However, the calculation of $q({\bf x}_{t-1} | {\bf x}_{t})$ depends on the entire dataset and is not straightforward. To address this, a neural network model $p_\theta$ is used to approximate these conditional probabilities with a Gaussian model:
\begin{equation}
p_\theta({\bf x}_{t-1} | {\bf x}_{t}) = \mathcal{N} ({\bf x}_{t-1} ; \mu_{\theta}({\bf x}_{t}, t), \sigma^2 {\bf I})  
\end{equation}
where $t$ is a timestep uniformly sampled from $1,2,\ldots,T$.
Then, $\mu_{\theta}({\bf x}_{t}, t)$ can be rewritten with noise prediction $\epsilon_\theta({\bf x}_t,t)$ as:
\begin{equation}
\mu_{\theta}({\bf x}_{t}, t) = \frac{1}{\sqrt{\alpha}_t} \left({\bf x}_t - \frac{1- \alpha_t} {\sqrt{1 - \Bar{\alpha}_t}}       \epsilon_\theta({\bf x}_t,t) \right)
\end{equation}
where ${\alpha}_t = 1 - {\beta}_t$ and ${\bf x}_t = \sqrt{\Bar{\alpha}_t} {\bf x}_0 - \sqrt{1- \Bar{\alpha}_t} \epsilon$ with $\Bar{\alpha}_t = \prod_{t=1}^t \alpha $. To learn to estimate $\epsilon_\theta({\bf x}_t,t)$ between consecutive samples ${\bf x}_{t-1}$ and ${\bf x}_t$, the training loss is defined as follows: 
\begin{equation}
L = \mathbb{E}_{t,{\bf x}_0,\epsilon} || \epsilon - \epsilon_\theta({\bf x}_t,t) ||^2_2
\label{eq:loss}
\end{equation}

\noindent {\bf UniDiffuser } UniDiffuser \cite{bao2022one} introduced an approach to handle multi-modal data distributions in the diffusion process in a unified manner. It defines the conditional expectations in a general form, $ \mathbb{E} [\epsilon_x, \epsilon_y | {\bf x}_{t^x},  {\bf y}_{t^y}]$, for all $0 \leq t^x, t^y \leq T$, where $t^x$ and $t^y$ represent two potentially different timesteps. ${\bf x}_{t^x}$ and ${\bf y}_{t^y}$ are the corresponding perturbed data. With this formulation, marginal diffusion, conditional diffusion and joint diffusion can be achieved by setting $t^y = T$, $t^y=0$ and $t^x = t^y = t$, respectively. To this end, a joint noise prediction network $\epsilon_\theta({\bf x}_{t^x}, {\bf y}_{t^y}, t^x,t^y)$ is employed to predict noise $\epsilon_\theta =[\epsilon_\theta^x, \epsilon_\theta^y]$ injected into ${\bf x}_{t^x}$ and ${\bf y}_{t^y}$, which is trained by minimizing the following loss:
\begin{equation}
\mathcal{L}_{\rm uni} =\mathbb{E}_{{\bf x}_0,{\bf y}_0, \epsilon_x, \epsilon_y, {\bf x}_{t^x}, {\bf y}_{t^y} }|| [\epsilon^x,\epsilon^y ] -  \epsilon_\theta({\bf x}_{t^x}, {\bf y}_{t^y}, t^x,t^y) ||^2_2
 \nonumber
\label{eq:loss_joint}
\end{equation}
where ${\bf x}_0$ and  ${\bf y}_0$ are the data points, $\epsilon_x$ and $\epsilon_y$ are sampled from Gaussian distributions, and $t^x$ and $t^y$ are independently and uniformly sampled from the range $1,2,\ldots,T$. Furthermore, UniDiffuser is directly applicable to classifier-free guidance (CFG), which is a method introduced to enhance the sample quality of conditional diffusion models \cite{Ho2022ClassifierFreeDG}, without modifying the training loss. For instance, ${\bf x}_0$ conditioned on ${\bf y}_0$ can be generated as follows:
\begin{equation}
\hat{\epsilon}_\theta^x({\bf x}_t, {\bf y}_0, t)  = (1+s_{\rm g})\epsilon_\theta^x({\bf x}_t, {\bf y}_0, t,0) - s_{\rm g} \epsilon_\theta^x({\bf x}_t, t,T) 
\label{eq:cfg}
\end{equation}
where $s_{\rm g}$ is a guidance scale and $\epsilon_\theta^x({\bf x}_t, {\bf y}_0, t,0)$ and $ \epsilon_\theta^x({\bf x}_t, t,T) $ are the conditional and unconditional models, respectively. 

\noindent {\bf Score Distillation Sampling (SDS) } A loss calculation framework called Score Distillation Sampling (SDS) is proposed by DreamFusion \cite{poole2022dreamfusion} for utilizing pre-trained diffusion models in optimizing and regularizing a neural 3D scene model. The scene model is defined by a parametric function of the form $x = g(\phi)$ capable of generating an image $x$ from the desired camera pose. In this context, $g$ is a volumetric renderer such as NeRFs and $\phi$ is a Multi-Layer Perceptron (MLP) that models a 3D volume. Given a text condition ${\bf y}_0$, SDS derives the gradient to update $\phi$ in such a way that:
\begin{equation}
\label{eq:SDS}
\nabla_\phi \mathcal{L}_{\rm SDS}(\phi, g(\theta)) = \mathbb{E}_{t,\epsilon} \left[\omega(t)(\epsilon_\theta({\bf x}_t, {\bf y}_0, t) -\epsilon) \frac{\partial \phi}{\partial x}  \right]   \nonumber
\end{equation}
where $\omega(t)$ is a weighting function. In practice, the conditional noise prediction model  $\epsilon_\theta({\bf x}_t, {\bf y}_0, t)$ is replaced with the classifier free guidance one, $\hat{\epsilon_\theta}({\bf x}_t, {\bf y}_0, t)$.

\begin{figure}[t]
\begin{center}
 \includegraphics[width=0.95\linewidth]{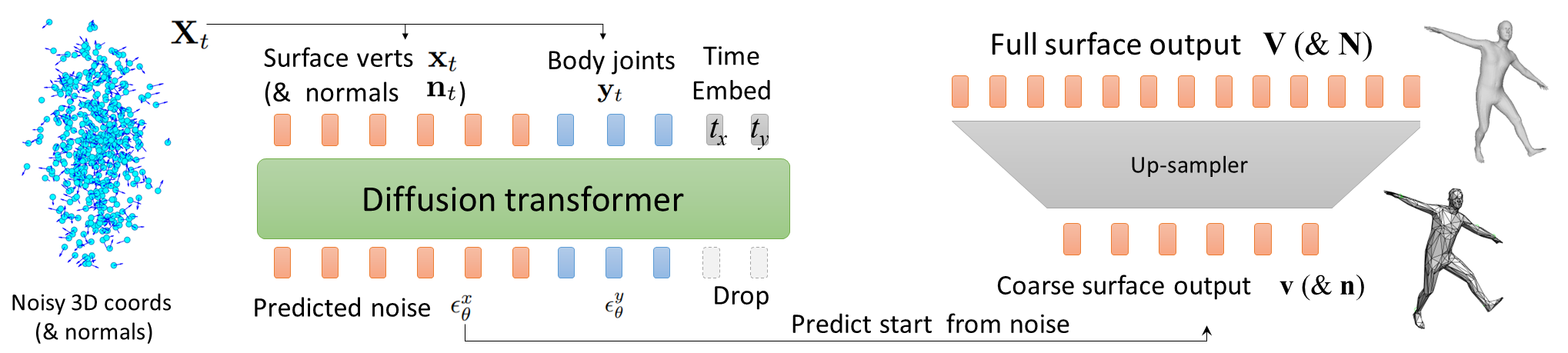}
 \vspace{-10pt}
 \caption{Overview. DiffSurf consists of a diffusion transformer and an up-sampler. The diffusion transformer takes in the noisy 3D coordinates of surface vertices ${\bf x}_t \in  \mathbb{R}^{N \times 3}$ and body joints ${\bf y}_t \in \mathbb{R}^{J \times 3}$. It processes these two modalities of data along with their corresponding timestep tokens $t_x$ and $t_y$. The transformer then outputs noise predictions for vertex and joint tokens, $\epsilon_\theta^x$ and $\epsilon_\theta^y$, respectively. For the 3D surface generation of man-made objects, we also input the noisy surface normals ${\bf n}_t \in  \mathbb{R}^{N \times 3}$ corresponding to vertex tokens into the diffusion transformer. Once the 3D coordinates of surface vertices ${\bf v}$ (and normals ${\bf n}$) are generated, up-sampling is optionally performed to obtain the full dense surface output ${\bf V}$ (and  ${\bf N}$). } 
 \vspace{-20pt}
 \label{fig:overview}
\end{center}
\end{figure}

\section{Method}

We introduce DiffSurf, a general network architecture for generating, editing and reconstructing 3D surfaces based on a plain diffusion transformer model, which is extendable to a wide range of object types. In addition, we introduce downstream methodologies to leverage pre-trained DiffSurf models for solving various 3D processing tasks.

For the generation of posed 3D surfaces, we propose to incorporate the vertex-based mesh recovery paradigm \cite{lin2021end-to-end,lin2021-mesh-graphormer,cho_arxiv.2207.13820} from human mesh recovery into 3D shape generation. Then, point-to-point correspondences between shapes are ensured by learning from training meshes with the same connectivity. This is a straightforward yet effective strategy, which is overlooked by the recent 3D shape generation literature, when unconditionally generating human and animal 3D surfaces in different poses where correspondences across shapes are vital. To the best of our knowledge, DiffSurf is the first 3D diffusion model capable of unconditionally generating 3D surfaces of articulated objects in diverse poses.

 \begin{wrapfigure}[6]{r}[0mm]{50mm}
\centering
\vspace{-10pt}
\includegraphics[keepaspectratio,width=50mm]{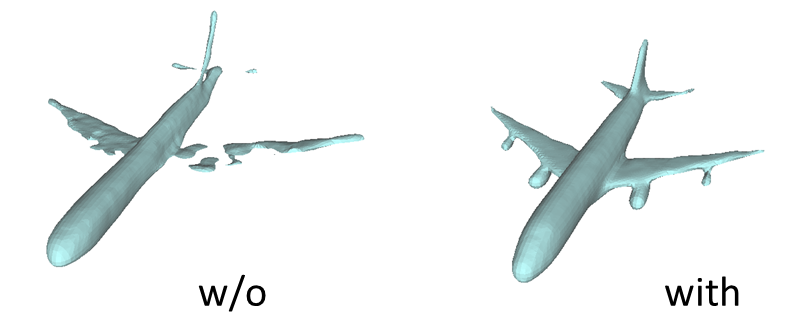}
\vspace{-20pt}
\caption{w/o and with surface normals.}
\label{fig:withnormal} 
\end{wrapfigure}

As for generating man-made objects, we propose a strategy for capturing better geometry by incorporating surface normals into the diffusion process. This approach not only leads to better generation results but also provides a more informative input for the subsequent surface reconstruction post-processing \cite{Peng2021SAP, zeng2022lion, slide2023} than using point sets alone, as shown in Fig. \ref{fig:withnormal}. This is particularly useful for generating 3D surfaces with complex geometry and topological differences, although point-to-point correspondences may be lost in this case. Unlike the recent diffusion models based on implicit functions  \cite{mo2023dit3d,cheng2023sdfusion}, in the diffusion process, we avoid using a volumetric representation whose complexity grows cubically w.r.t voxel resolution. Instead, we directly work on an explicit point representation through the diffusion transformer to exploit long-range dependencies between points. As a result, DiffSurf only needs a single diffusion model as opposed to previous hierarchical latent diffusion models which rely on two  diffusion models \cite{zeng2022lion}, making our model computationally more efficient.

\subsection{Network architecture}

We describe our diffusion transformer architecture for generating 3D surfaces. It draws inspiration from vertex-based human mesh recovery approaches \cite{lin2021end-to-end,lin2021-mesh-graphormer,cho_arxiv.2207.13820} and point cloud latent diffusion models \cite{luo2021diffusion, zeng2022lion, slide2023}, which process explicit 3D point-based representations in neural network models. Specifically, we have designed a UniDiffuser-like multi-modal diffusion transformer architecture \cite{Xu_2023_ICCV} that predicts noise from noisy 3D coordinates of surface vertices and body joints, treating them as two distinct modalities. The incorporation of body joints not only facilitates more effective training \cite{lin2021end-to-end} but also provides landmark controls  \cite{slide2023} for manipulating shape and pose. Consequently, we introduce a simple yet versatile transformer architecture for generating and reconstructing 3D surfaces of articulated objects in various shapes and poses, thereby enabling various downstream 3D processing tasks as illustrated in Fig. \ref{fig:overview}. DiffSurf consists of 1) a diffusion transformer and 2) a mesh up-sampler.


\noindent{\bf Diffusion transformer blocks } The inputs to the diffusion transformer consist of noisy 3D coordinates for a set of joint query tokens $Q_{\rm J}=\{ Q^1_{\rm J} \ldots Q^J_{\rm J} \} $ and coarse vertex query tokens $Q_{\rm V}= \{ Q^1_{\rm V} \ldots Q^N_{\rm V} \} $, corresponding to an articulated body mesh comprising $J$ joints and $N$ vertices. The input noisy 3D coordinates of surface vertices, body joints and their concatenations are respectively denoted as ${\bf x}_t \in \mathbb{R}^{N \times 3}$, ${\bf y}_t \in \mathbb{R}^{J \times 3}$ and ${\bf X}_t \in  \mathbb{R}^{(J + N) \times 3} $. The diffusion transformer processes these two modalities of data and their corresponding timesteps $t_x$ and $t_y$ as tokens. It outputs noise predictions for vertices and joints, $\epsilon_\theta^x$ and $\epsilon_\theta^y$. For the generation of man-made objects, we concatenate the noisy 3D coordinates of vertices ${\bf x}_t$ with the corresponding surface normals ${\bf n}_t \in \mathbb{R}^{N \times 3}$ to construct an ${N \times 6}$ matrix, which is then input to the diffusion transformer. Our diffusion transformer consists of 7 layers of transformer blocks and input/output MLP layers. Each transformer block has hidden layers with the dimension of 256 channels. The input MLP layer converts ${\bf x}_t$ and ${\bf y}_t$ into 256-dimensional embedding features and the output MLP layer converts the features processed by transformer into $\epsilon_\theta^x$ and $\epsilon_\theta^y$.

\noindent{\bf Up-sampling } After a coarse surface comprising $N$ vertices, ${\bf v} \in \mathbb{R}^{N \times 3} $ and corresponding surface normals ${\bf n} \in \mathbb{R}^{N \times 3} $ are produced from the noise prediction $\epsilon^x_\theta$ by the diffusion transformer, we  apply an upsampling operation. For human and animal generation, where point-to-point correspondences i.e. mesh connectivity is fixed, an upsampling technique based on MLPs similar to \cite{lin2021end-to-end,lin2021-mesh-graphormer,yoshiyasu2023-deformer} is adopted to obtain a dense mesh (see Fig. \ref{fig:overview}) with $M$ vertices, ${\bf V} \in \mathbb{R}^{M \times 3} $. For the surface generation of man-made objects, refinement and upsampling based on the improved PointNet++ model \cite{lyu2022conditional} are applied to increase the number of points and normals by a factor of $\times5$ \cite{slide2023}. Subsequently, a learning-based surface reconstruction technique called Shape-As-Points (SAP) \cite{Peng2021SAP} is employed to convert the upsampled points ${\bf V}$ and normals ${\bf N}$ into a mesh.

\subsection{Downstream methodologies}

Here, we demonstrate that DiffSurf is capable of performing a series of downstream tasks in 3D surface editing and reconstruction.   


\noindent {\bf Pose conditioned mesh generation } DiffSurf can generate a mesh that is conditioned on 3D skeleton landmark locations, such as those obtained using motion capture and image-based 3D pose regressors. Essentially, this process of conditional mesh generation involves feeding 3D body joint locations as queries into the diffusion transformer and setting the timestep to $t_y=0$. Naively feeding 3D joint locations into DiffSurf results in slight discrepancies between the mesh and the joints. To address this, we leverage CFG (Eq. (\ref{eq:cfg})) to push the mesh toward joint locations and improve alignments between them. We found that setting the CFG weight to around $s_{\rm g} = 1$ effectively improves alignment while preserving the mesh structure. Excessively increased CFG weights, e.g., $s_{\rm g} > 3$, can result in distortion of the mesh (as shown in the \yy{Appendix}).

\noindent {\bf Body shape variation } By feeding a skeleton to DiffSurf as a condition and performing sampling with varying random noise, we can generate meshes in different body shapes, as shown in Fig. \ref{fig:teaser} (right). However, this approach does not allow for changes in body heights and segment lengths. To address this, we adopt a two-step strategy. The first step involves unimodal generation to create a batch of skeletons with different body poses and styles. The second step then adjusts the segment lengths of one of generated skeletons based on others in the batch. By inputting these modified skeletons into DiffSurf, we can generate meshes in diverse body styles while maintaining the pose.   

\noindent {\bf Shape morphing } DiffSurf is capable of morphing between two meshes with different poses and shapes. This is achieved by blending two meshes represented as the Gaussian noise, ${\bf x}^1_T$ and ${\bf x}^2_T$, through spherical linear interpolation (SLERP) \cite{slerp},  $\hat{\bf x}_T = {\rm SLERP}({\bf x}^1_T, {\bf x}^2_T, w )$, where $w$ is an interpolation weight within the range $[0,1]$. Setting $w$ outside this range $[0,1]$ e.g., $[-0.25,1.25]$, results in extrapolation. Given the new noise $\hat{\bf x}_T$, a mesh is then sampled from it using DiffSurf. It is noteworthy that DiffSurf has the capability to simultaneously handle both shape and pose variations. 

\noindent {\bf Shape refinement } Instead of starting from Gaussian noise to generate a mesh, DiffSurf can refine a mesh exhibiting noise and distortions by leveraging the SDS gradients as calculated in Eq. (\ref{eq:SDS}). Analogous to mesh fairing \cite{desbrun99}, a sequence of refined meshes can be constructed by explicitly applying the SDS gradients to a mesh: 
\begin{equation}
{\bf X}^{l+1}  = {\bf X}^{l} - \nabla_\phi \mathcal{L}_{\rm SDS}(\phi,{\bf X}^{l}) 
\end{equation}
It should be noted that, in contrast to DreamFusion \cite{poole2022dreamfusion},  ${\bf X}^{l}$ is not an output from a neural network model but rather the 3D coordinates of surface vertices and body joints.

\noindent {\bf Control point deformation } DiffSurf facilitates data-driven mesh deformation, similar to prior shape editing techniques \cite{SumnerZGP05, Frohlich2021, Gao2021}, by creating a mesh in a plausible shape and pose through the specification of a set of control points and fitting the mesh toward them. Building upon \cite{poole2022dreamfusion}, we devise a loss function derived from differential mesh properties of the SDS target mesh to preserve local geometries of the mesh. We minimize the following total loss $L_{\rm def}$ to optimize ${\bf X}$: 
\begin{align}
L_{\rm def} ({\bf X}) &= L_{\rm SDS} + L_{\rm SDS}^{\rm edge} + L_{\rm SDS}^{\rm lap}  + L_{\rm consist}  + L_{\rm CP}
\label{eq:deform}
\end{align}
where $L_{\rm SDS}$, $L_{\rm SDS}^{\rm edge}$ and $L_{\rm SDS}^{\rm lap}$ represent losses defined by the distances between the SDS targets and predictions for the 3D vertex coordinates, edges and Laplacian coordinates of the coarse mesh, respectively. $L_{\rm consist}$ maintains the consistency between the joint and mesh predictions, defined by the distances between the optimized joints and the regressed joints, which are calculated from the coarse mesh vertices using the joint regressor. The regressed joints ${\bf j}_{\rm reg} \in \mathbb{R}^{J  \times 3}$ are calculated from the mesh vertices using the joint regressor, ${\bf j}_{\rm reg} = \cal{J}{\bf V}$, where ${\cal J}  \in \mathbb{R}^{J \times M} $ is a joint regressor matrix  \cite{SMPL:2015,Zuffi:CVPR:2017,MANO:SIGGRAPHASIA:2017}. $L_{\rm CP}$ quantifies the distances between the optimized joint locations and the control points (see the \yy{Appendix} for more details).

Figure \ref{fig:control_points} (left) shows the comparisons of the loss terms. Using $L_{\rm SDS}$ and $L_{\rm CP}$, DiffSurf is able to fit a human mesh towards control points, but there remain some distances between them. Incorporating $L_{\rm consist}$ improves the fit but introduces distortions around the control points. Adding $L_{\rm SDS}^{\rm edge}$ and $L_{\rm SDS}^{\rm lap}$ remedies this issue by considering the differential properties of the coarse mesh to preserve its local geometry.

\begin{figure}[t]
\begin{center}
 \includegraphics[width= 0.95 \linewidth]{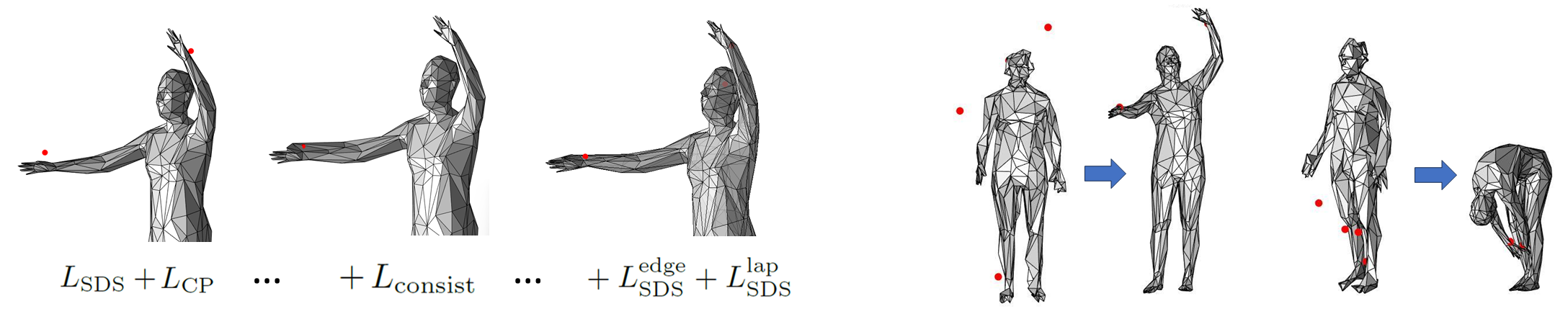}
 \vspace{-10pt}
 \caption{Control point deformation. Left: comparison of the loss terms. Right: deformation examples obtained using five control points (head, wrists and ankles, marked in red). The process starts with the rest pose and a human mesh is deformed to align with the control points. DiffSurf can deform a mesh into extremely different poses, such as a forward-bending posture. }
\vspace{-20pt}
 \label{fig:control_points}
\end{center}
\end{figure}



\noindent {\bf 2D keypoint fitting }
While previous research \cite{bogo2016keep,kolotouros2021prohmr} has addressed the challenge of fitting a parametric model to 2D landmarks, we propose an alternative vertex-based fitting approach for this task. Consequently, our method is applicable to both parametric and vertex-based mesh recovery approaches and improves their mesh recovery results. 

Starting from the initial solution derived from a mesh recovery approach, DiffSurf optimizes mesh vertices and body joints to improve their alignment with 2D keypoint locations. The loss function for 2D keypoint fitting is defined by modifying Eq. (\ref{eq:deform}) slightly as follows:
\begin{equation}
L_{\rm fit} ({\bf X}) = L_{\rm SDS} + L_{\rm SDS}^{\rm edge} + L_{\rm SDS}^{\rm lap}  + L_{\rm consist}  + L_{\rm 2D}
\label{eq:fit}
\end{equation}
where $L_{\rm 2D}$ measures the discrepancies between the ground truth and predicted 2D keypoints. These 2D keypoint predictions are obtained by projecting 3D joint positions using the camera parameter predictions from the mesh recovery approach. In addition, as DiffSurf is trained on the dataset with global position and rotation aligned at the root, we rigidly align the mesh prediction with the canonical orientation before subtracting the SDS gradients, such that: $T({\bf X} - \nabla_\phi \mathcal{L}_{\rm SDS}(\phi, T^{-1}({\bf X}) )$, where $T$ is the global transformation of the mesh recovery result w.r.t the canonical orientation. The global rotation can be the root orientation predicted by the mesh recovery approach  when the global pose is available. Otherwise, a coordinate frame defined from the body joint predictions can be used to obtain $T$ e.g., the x-axis and y-axis are defined from the unit vectors emanating from the pelvis to the neck and from the right hip to the left hip.   


\noindent {\bf Mesh generation from 3D keypoints } DiffSurf is capable of reconstructing a 3D mesh from 3D joint locations by employing the pre-trained DiffSurf model for pose-conditioned surface generation. To achieve this, we first predict 3D body joint locations from an image using a 3D pose regressor, which produces the 3D positions of 14 body joints. Subsequently, these 3D joint positions are inputted into DiffSurf to perform conditional mesh generation with Eq. (\ref{eq:cfg}). Similar to the SDS-based 2D keypoint fitting, this approach requires aligning the mesh with the canonical orientation prior to mesh sampling. It is noteworthy that this modular human mesh recovery design, based on DiffSurf, decouples an image-based 3D pose regressor from a mesh generator, enabling its training even when image-mesh paired data are not available. The architecture of our 3D pose regressor used here is transformer-based (see the \yy{Appendix}).

\section{Experiments}

\subsection{Dataset and metrics}

\noindent {\bf 3D generation } We  trained our DiffSurf models separately on publicly available 3D datasets: SURREAL \cite{varol17_surreal}, AMASS \cite{AMASS:2019}, FreiHAND \cite{Freihand2019}, BARC \cite{BARC:2022}, Animal3D \cite{xu2023animal3d} and ShapeNet \cite{chang2015shapenet}. We follow 3D-CODED \cite{groueix2018b} for the SURREAL train/test split definition. The global positions of meshes used in the training are aligned at the root position and oriented to face forward. The body joints are obtained from meshes using joint regressors \cite{SMPL:2015,Zuffi:CVPR:2017} for humans and animals. For ShapeNet objects, we feed sparse latent points generated by SLIDE \cite{slide2023} as body joint tokens to transformer. 

%
Evaluation of 3D human generation was conducted on the SURREAL testset (200 meshes) and DFAUST \cite{dfaust2017} (800 meshes). The standard metric used for evaluating 3D generation is the 1-NNA metric \cite{pointflow}, which quantifies the distributional similarity between generated shapes and the validation set. This metric assesses both the quality and diversity of the generated results. For human generation, given the differences in global orientations between validation and training meshes, we first perform a rigid alignment of the predicted mesh with the validation meshes before calculating the 1-NNA metric. We refer to this modified metric as Rigid Aligned 1-NNA (RA-1-NNA). 


\begin{figure}[t]
\begin{center}
 \includegraphics[width=0.95\linewidth]{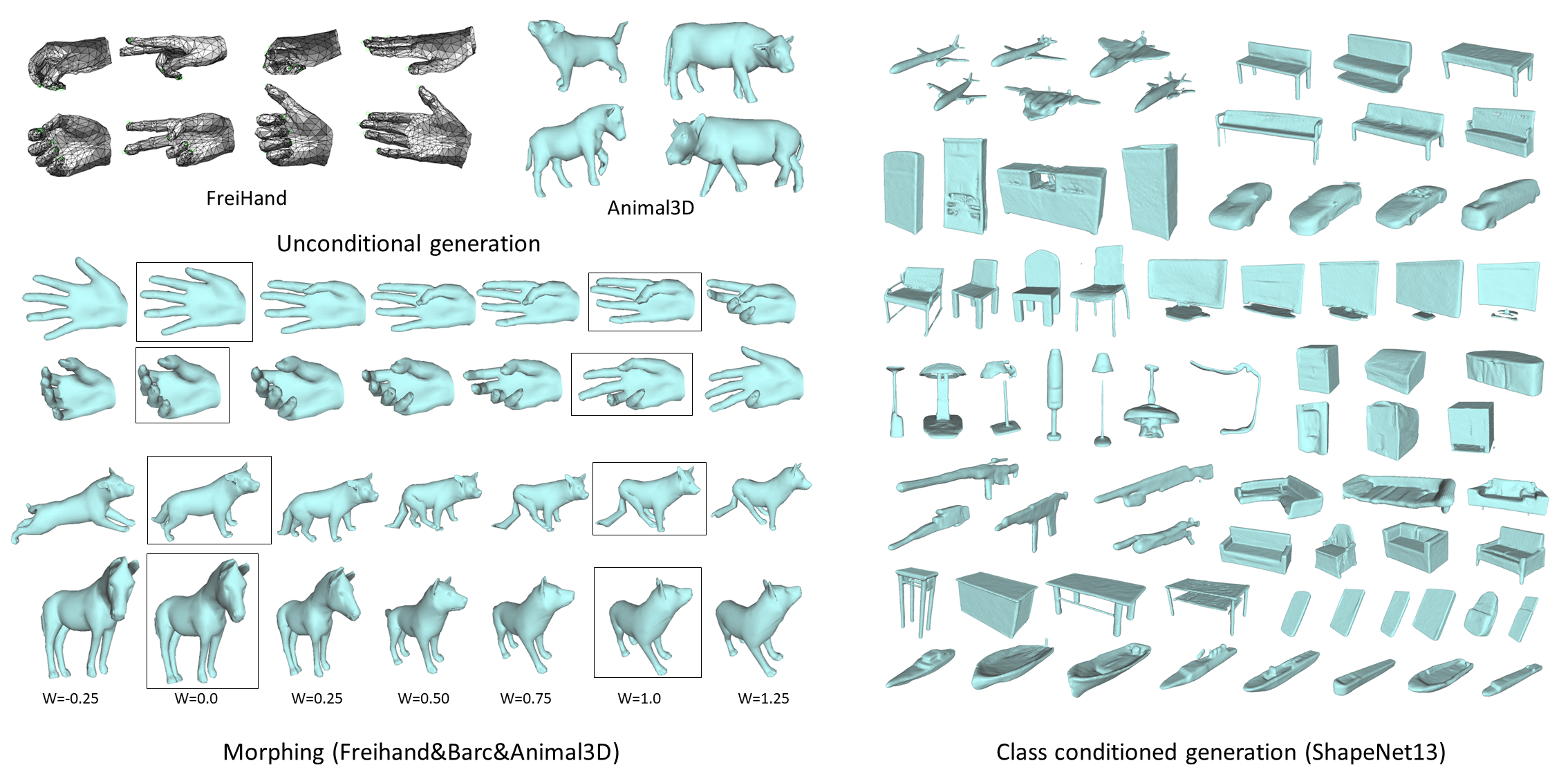}
 \vspace{-10pt}
 \caption{Left: Unconditional generation and morphing of hands, dogs and animals. Right: class conditioned generation of man-made objects.  }
 \vspace{-25pt}
 \label{fig:hands}
\end{center}
\end{figure}

\noindent {\bf Human mesh recovery } We trained our 3D pose regressor using publicly available datasets, adopting the mixed dataset training strategies as outlined in \cite{lin2021end-to-end,Kocabas_PARE_2021}. The datasets used include Human3.6M \cite{h36m_pami}, MPI-INF-3DHP \cite{mono-3dhp2017}, COCO \cite{LinMBHPRDZ14}, MPII \cite{andriluka14cvpr} and LSPET \cite{Johnson2011}. For training, we utilized the 3D joint labels from Human 3.6M and 2D keypoint labels from all the dataset. Additionally, we conduct another training experiment using 3D body joint labels obtained from pseudo 3D meshes produced by EFT \cite{joo2020eft} on the in-the-wild image datasets. Unlike recent mesh transformer approaches \cite{lin2021end-to-end,cho_arxiv.2207.13820}, we did not use 3DPW \cite{vonMarcard2018} as training data for fine-tuning on 3DPW, but instead only performed evaluations on its test set. 

We used the following three standard metrics for evaluation: MPJPE, PA-MPJPE and MPVE. Mean-Per-Joint-Position-Error (MPJPE) measures the Euclidean distances between the ground truth and the predicted joints. The PA-MPJPE metric, where PA stands for Procrustes Analysis,  measures the error of the reconstruction after removing the effects of scale and rotation.  Mean-Per-Vertex-Error (MPVE)  measures the Euclidean distances between the ground truth and the predicted vertices.

\subsection{Training and sampling}
The training of DiffSurf involves two steps: training of the diffusion model and the up-samplers are done separately. We use pre-trained up-sampler models for fixed and varied topology cases (see the \yy{Appendix} for the details on the network architectures and their training). Our diffusion transformer model is trained with a batch size of 256 for 400 epochs on 4 NVIDIA V100 GPUs for the SURREAL dataset, and for 200 epochs on 8 NVIDIA A100 GPUs for the AMASS dataset. It takes about 1 day for both cases. For the BARC, Animal3D and ShapeNet objects, we extend the training of diffusion transformer to 4000-8000 epochs because they contain fewer meshes than SURREAL and AMASS. The dataset statistics are provided in \yy{Appendix}. We use the Adam optimizer for training our models, while reducing the learning rate by a factor of 10 after $1/2$ of the total training epochs beginning from $1 \times 10^{-4}$. For the training objective of DiffSurf, we adopt the v-prediction parameterization \cite{salimans2022progressive,lucidrain} and employ the DDIM \cite{song2020denoising} sampler along with a sigmoid variance scheduler. We set the diffusion time step to $T =1000$ and tested sampling steps in the range [1-250].

\begin{figure}[t]
\begin{center}
 \includegraphics[width=0.95\linewidth]{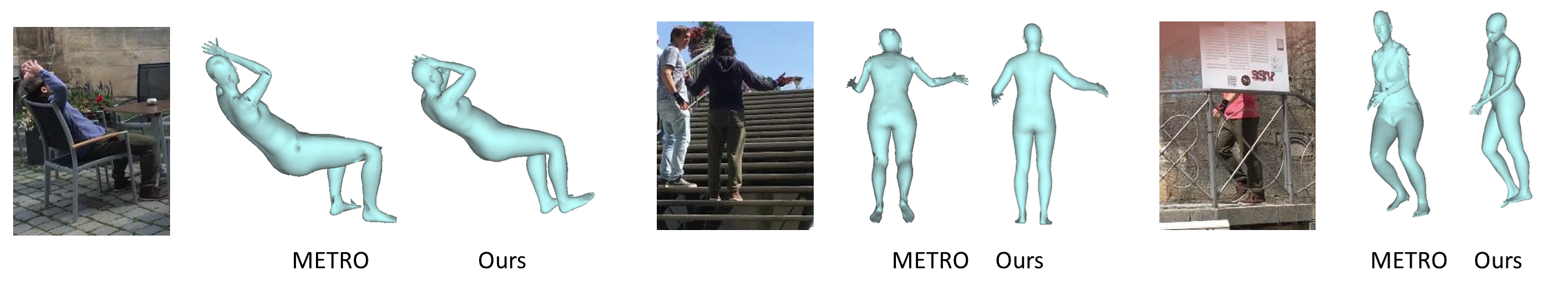}
 \vspace{-10pt}
 \caption{Example results of human mesh recovery by DiffSurf on the 3DPW dataset. Compared to METRO, DiffSurf produces less distorted results, especially in occluded situations. }
 \vspace{-30pt}
 \label{fig:qulalitativehmr}
\end{center}
\end{figure}

\subsection{Downstream applications}
\vspace{-10pt}

\noindent {\bf Unconditional shape generation } Figures \ref{fig:teaser} (left) and \ref{fig:hands} (left) depict unconditional 3D mesh generation results of humans, hands, dogs, mammals and man-made objects based on DiffSurf. 
DiffSurf trained on FreiHAND can generate 3D hand meshes in a variety of poses, including thumbs up, victory (peace), open and close. The results obtained using the BARC and Animal3D dataset indicate that DiffSurf can handle a range of dog breeds, from small to large, as well as different species. Our approach can achieve class-conditioned generation of ShapeNet 13 objects by inputting class labels to the transformer as in U-ViT, which includes generation of topologically different shapes such as the lamp examples. These results demonstrate the ability of DiffSurf to generate 3D meshes in diverse shapes and poses.




\noindent {\bf Pose conditioned generation and body shape variations } Figure \ref{fig:teaser} and the \yy{Appendix} demonstrate the outcomes of pose-conditional mesh generation and body shape variation using DiffSurf. When provided with different 3D joint locations while sampling from the same mesh noise input, DiffSurf can generate various poses of the same body mesh. Variations in body shape are realized by altering the mesh noise input.

\begin{figure}[t]
\begin{center}
 \includegraphics[width=0.95\linewidth]{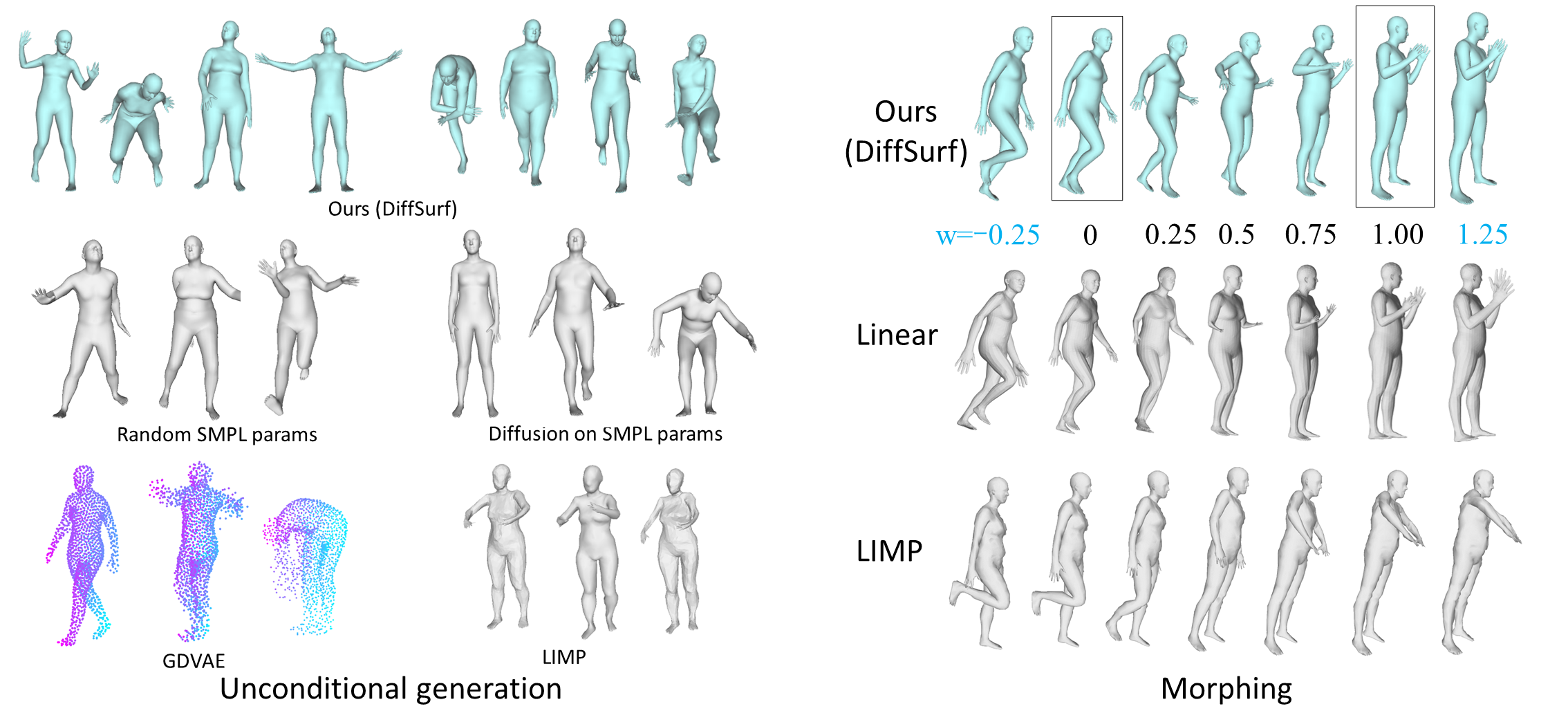}
 \vspace{-10pt}
 \caption{Qualitative comparisons with previous techniques are presented. Left: comparison of unconditional generation results.  Right: comparison of morphing results against linear interpolation in 3D Euclidean space and LIMP. }
  \vspace{-20pt}
 \label{fig:qualitative3d}
\end{center}

\end{figure}

\noindent {\bf Shape morphing } In Figs. \ref{fig:teaser}, \ref{fig:hands} (left),  \ref{fig:qualitative3d} (right)  and the \yy{Appendix}, we present morphing results produced by DiffSurf, where two meshes are interpolated in the Gaussian noise space. In contrast to linear interpolation of 3D vertex coordinates in the 3D Euclidean space, which results in a straight-line interpolation trajectory leading to the artifacts such as arm shrinkage and hand expansions, DiffSurf yields visually plausible interpolation outcomes (see Fig. \ref{fig:qualitative3d} right). Since LIMP is an approach that learns from a small amount of meshes (in static poses), its pose representation capability is limited. As shown in Fig. \ref{fig:hands} (left), this approach can morph between two objects with different shapes, such as horse and dog, which showcases the ability of DiffSurf for handling body shapes and poses together and its potential for becoming a viable alternative to prior nonlinear and data-driven morphing techniques \cite{Frohlich2021,Gao2021}.


\noindent {\bf Control point deformation } Figure \ref{fig:control_points} (right) illustrates the results of control point deformation. In these examples, five control points are specified at the head, wrists and ankles and the deformation begins with the rest pose. Through gradient-based optimization and progressively decreasing diffusion time steps, the mesh is refined to a pose that conforms to the control points without distortions. 

\begin{table}[tb]
\centering
\caption{Comparisons with baseline models on unconditional human generation. The RA-1-NNA metric [\%] assesses the diversity and quality of generated results. A lower value on this metric signifies superior performance. SD indicates the  standard deviation. }
\begin{minipage}[c]{0.45\textwidth}
\vspace{-5pt}
\scalebox{0.7}[0.7]{
\begin{tabular}{c c c}
\hline 
 & SURREAL & DFAUST \\
\hline 
Random SMPL (SD$\times 0.2$) & 81.1 & 92.8 \\
Random SMPL (SD$\times 1.0$) & 67.2 & 71.4 \\
Random SMPL (SD$\times 3.0$) & 71.6 & 95.4 \\
VPoser \cite{SMPL-X:2019}  & 60.7 & 70.2 \\
Parametric Diffusion  & 59.6 & 76.2 \\
\hline 
\end{tabular}
}
\subcaption*{*Trained on AMASS}
\end{minipage}
\begin{minipage}[c]{0.45\textwidth}
\vspace{-20pt}
\scalebox{0.8}[0.8]{
\begin{tabular}{c | c | c c}
\hline 
& \diagbox{Train}{Test} & SURREAL & DFAUST \\
\hline 
GDVAE \cite{9010824} & SURREAL &93.8& 98.1\\
LIMP \cite{LIMP2020} & FAUST & 81.3 & 93.3\\
DiffSurf & SURREAL & 54.4 &69.6\\
DiffSurf & AMASS & {\bf 54.0} &  {\bf 69.5} \\
\hline 
\end{tabular}
}
\end{minipage}
\label{tab:generation}
\vspace{-30pt}
\end{table}

\noindent {\bf Human mesh recovery from image } Figure \ref{fig:qulalitativehmr} visualizes the mesh recovery results on the 3DPW dataset obtained by DiffSurf. Even though DiffSurf is trained without image-mesh paired data, it produces visually pleasing results without noticeable artifacts.

\subsection{Comparisons}

\noindent {\bf Unconditional human mesh generation } Here, DiffSurf is compared against five baselines. We employ parametric baseline approaches: Random SMPL which draws SMPL body shape/pose parameters randomly from $\times [0.2,1.0, 3.0]$ the standard deviations of AMASS parameter collections to generate human body meshes; VPoser \cite{SMPL-X:2019} which learns pose priors with VAEs; parametric diffusion transformer that generates SMPL parameters. We also compared DiffSurf with the previous generative models for surfaces based on VAEs: GDVAE \cite{9010824} and LIMP \cite{LIMP2020}. 

Figure \ref{fig:qualitative3d} presents a qualitative comparison, while Table \ref{tab:generation} provides quantitative comparisons using the RA-1-NNA metric. As shown in Table \ref{tab:generation}, naively producing body meshes from random SMPL parameters proved to be far inferior to our approach. DiffSurf also outperforms VPoser and parametric diffusion, which use rotational parametrization of pose that is usually difficult to learn with neural networks. Since GDVAE relies on a point cloud representation, the results exhibit outliers especially around hands and feet (see Fig. \ref{fig:qualitative3d}). LIMP is based on mesh representation and preserves mesh structure by maintaining both extrinsic and intrinsic surface properties. However, LIMP's diversity in body poses and shapes appears to be constrained, likely due to its training strategy relying on a limited dataset. As both methods are based on MLP-based VAEs, their generated sample quality is not as high as that produced by DiffSurf.

\noindent {\bf Human mesh recovery from image } Table \ref{tab:SOTA} presents a comparison of our method with previous human mesh recovery approaches, which are  divided into parametric and vertex-based categories, on the 3DPW and Human3.6M datasets. Note that none of methods used the 3DPW dataset in training. DiffSurf achieves top-level performances among vertex-based approaches. Our method is also comparable to recent diffusion based methods \cite{dat2023,cho2023generative}, even though ours is not trained end-to-end on image-mesh paired dataset. Since DiffSurf does not explicitly relate its generation to an image, performance is affected by random input mesh noise $\epsilon^x_\theta$ that possibly alters body styles and twisting joint angles of generations. We show in the \yy{Appendix} how multiple hypotheses on the input mesh noise can further possibly improve DiffSurf's performance.

\noindent {\bf 3D Human mesh fitting to 2D keypoints } Table \ref{tab:fitting} shows a comparison of optimization approaches that fits a mesh with ground truth (GT) 2D keypoints.  With our SDS-based approach, the MPVPE and PA-MPJPE errors for both parametric and vertex-based mesh recovery techniques decrease, outperforming the previous fitting approaches that utilized GT keypoints \cite{bogo2016keep,  kolotouros2021prohmr, joo2020eft, lgd2020}. In fact, PA-MPJPE dropped by approximately 2pts and 9pts for HMR-EFT \cite{joo2020eft} and METRO \cite{lin2021end-to-end}, respectively. These results suggest that DiffSurf can potentially aid in the creation of an image-mesh paired dataset with improved alignment between images and meshes.

\noindent
 \begin{minipage}[t]{0.55\textwidth}

\vspace{-5pt}
\captionof{table}{Comparisons with other 3D human mesh recovery approaches on 3DPW. No fine-tuning on 3DPW performed.}
    \scalebox{0.65}[0.65]{
    \begin{tabular}{c c c c c c}
\hline 
& \multirow{2}{*}{Method} & \multicolumn{2}{c}{3DPW} & \multicolumn{2}{c}{Human 3.6M}\\
&  & MPVE $\downarrow$ & PA-MPJPE $\downarrow$ & MPJPE $\downarrow$ & PA-MPJPE $\downarrow$ \\
\hline  
\multirow{6}{*}{\rotatebox{90}{Parametric}}  &SPIN \cite{kolotouros2019spin} & 116.4 & 59.2 & 62.5 & 41.1 \\
& ProHMR \cite{kolotouros2021prohmr}  & ---  & 59.8  & --- & 41.2 \\
& OCHHuman \cite{Khirodkar_2022_CVPR} & 107.1  & 58.3 & --- & --- \\
& DiffHMR \cite{luo2021diffusion} & 110.9 & 56.5  & --- & --- \\
& HMR-EFT \cite{joo2020eft} & ---  & 54.3 &---  & --- \\
& PARE \cite{Kocabas_PARE_2021} & {\bf 97.9} & {\bf 50.9} & 76.8 & 50.6 \\   
\hline  
\multirow{6}{*}{\rotatebox{90}{Vertex-based} }& METRO \cite{lin2021end-to-end} & 119.1 & 63.0 & 54.0 & 36.7 \\
& Pose2Mesh \cite{Choi_2020_ECCV_Pose2Mesh} & 106.3 & 58.3 & 64.9 & 46.3 \\
& GATOR \cite{10096870} & 104.5  & 56.8 & 64.0 & 44.7 \\
& HMDiff \cite{dat2023} & --- & ---  & \underline{49.3} & {\bf 32.4} \\
& DiffSurf  &  108.0 & 53.7  & {\bf 48.9} & \underline{36.1} \\
& DiffSurf-EFT  & \underline{102.6} & \underline{52.6} & 50.1 & 36.9  \\
\hline
\end{tabular}
}
\label{tab:SOTA}
\end{minipage}
\hspace{10pt}
\begin{minipage}[t]{0.4\textwidth}
  \centering
\captionof{table}{Comparisons with previous fitting approaches on 3DPW. GT 2D keypoints are used. }
\vspace{10pt}
  \scalebox{0.65}[0.65]{
\begin{tabular}{c c c }
\hline 
Method & MPVE $\downarrow$ &   PA-MPJPE $\downarrow$  \\
\hline 
SMPLify \cite{bogo2016keep} & ---  & 106.1 \\
LearnedGD \cite{lgd2020} & ---  & 55.9 \\
ProHMR + fitting \cite{kolotouros2021prohmr} & ---  & 55.1 \\
HMR-EFT + EFT fitting \cite{joo2020eft} & --- & 53.7 \\
HMR-EFT \cite{joo2020eft} + SDS fitting & 98.6   & 52.1  \\
METRO \cite{lin2021end-to-end} + SDS fitting & 103.0 & 54.2 \\
DiffSurf + SDS fitting & {\bf 93.7} & {\bf 48.7} \\
\hline 
\end{tabular}
}
\label{tab:fitting}
\end{minipage}
\vspace{-10pt}

\subsection{Ablation studies}

\noindent {\bf  Network architectures } To conduct the ablation study on our diffusion transformer model's components, we modified the elements within it and compared their performances. The basic network architecture from which we started is an adaptation of U-ViT \cite{bao2022all}, tailored to handle 3D mesh and body joint tokens (see Table \ref{tab:ablation}, top row). We investigated the impacts of long skip connections, different methods of incorporating time embedding, position embedding constructions and the effect of varying the network layer types (transformer or MLPs). As shown in Table \ref{tab:ablation}, the switch in the layer type from transformer to MLPs yields the most significant impact, indicating that the most critical component of DiffSurf is the transformer layer. As opposed to U-ViT \cite{bao2022all} for image generation, the use of long skip connection does not contribute to improving 3D human mesh generation. This may be attributed to the fact that the current form of the diffusion transformer in DiffSurf primarily processing coarse-level mesh vertices. Expressive human body generation that incorporates fine-grained details like finger poses and facial expressions could potentially benefit from long-skip connections.

\begin{table}[t]
\caption{Ablation studies: (a) Network components. Errors are measured for unconditional human generation using the RA-1-NNA metric on the SURREAL dataset; (b) Sampling time steps. Errors are measured for 3D human mesh recovery on the 3DPW dataset with a CFG weight $s_g =1.0$; (c) CFG scale factor $s_{\rm g}$. The error is measured by PA-MPJPE$\downarrow$ on the 3DPW and H3.6M datasets with 10 DDIM sampling time steps. }
\vspace{-20pt}
\begin{minipage}[c]{0.5\textwidth}
\begin{subtable}{\textwidth}
\centering
\caption{Ablation study on network components. }
\vspace{-10pt}
\scalebox{0.7}[0.7]{
\begin{tabular}{c c c c c}
\hline 
Layer & Pos emb & Time emb & Long skip & 1NNA $\downarrow$ \\
\hline 
Transformer &  Learned & Token & Yes & 55.6  \\
Transformer &  Learned & Token & {\bf No} & {\bf 54.4} \\
Transformer & Learned & {\bf Add} & Yes & 55.4  \\
Transformer &  {\bf 3D Template} & Token & Yes & 54.8 \\
{\bf MLP} &  Learned & Token & Yes & 92.6  \\
\hline 
\end{tabular}
\label{tab:ablation}
}
\end{subtable}
\end{minipage}
\begin{minipage}[c]{0.5\textwidth}
\begin{subtable}{\textwidth}
\centering
\caption{Ablation study on sampling time steps. }
\vspace{-10pt}
\scalebox{0.7}[0.7]{
\begin{tabular}{c c c c c c c c}
\hline 
& 1 & 3 &  5 & 10& 20 & 30 &  100 \\
\hline 
MPVE $\downarrow$ & 230.4  & 115.8 & 107.3  & {\bf 105.4} & 105.8  & 106.1  & 106.9  \\
PA-MPJPE $\downarrow$ & 143.9  & 60.2 & 55.6  & 54.3 & {\bf 54.1}  & 54.2  & 54.7 
\\
fps & 35.9  & 32.05 & 25.25 &  21.2 & 13.6 & 8.81 & 3.38 \\
\hline 
\end{tabular}
\label{tab:timesteps}
}
       \end{subtable}

       \begin{subtable}{\textwidth}
           \centering
       \caption{Ablation study on CFG scale factor. }
\vspace{-10pt}
\scalebox{0.8}[0.8]{
\begin{tabular}{c c c c c c }
\hline 
 & $s_{\rm g} = 0.0$ & 0.1 &  0.3 & 0.5 & 1.0  \\
\hline 
3DPW  & {\bf 52.6} & 52.6 & 52.9 & 53.3 & 54.3   \\
H3.6M  & 37.9 & 37.0  & 36.2 & {\bf 36.1} & 36.7   \\
\hline 
\end{tabular}
\label{tab:cfg}
}
\end{subtable}
\end{minipage}
\vspace{-20pt}
\end{table}

\noindent {\bf Sampling timesteps }  Table \ref{tab:timesteps} presents an ablation study on the sampling timesteps for 3D human mesh recovery from an image. It is observed that an increase in the sampling timesteps leads to a decrease in MPVPE and PA-MPJPE, reaching optimal performance at approximately 10-20 steps. We also measured the inference time with respect to sampling timesteps. In this configuration, DiffSurf operates at nearly real-time speed, approximately 20 FPS, when performing DDIM sampling with 10 steps. Furtheremore, the sampling speed of unconditional human generation with 431 vertices were 37,  9  and 1.5 fps for 10, 50 and 250 DDIM sampling steps, respectively. The sampling speed of man-made objects generation (2048 points) were 17.8,  3.5  and 1.8 fps for 10, 50 and 100 DDIM sampling steps, respectively, which is roughly $7\times $ faster than LION \cite{zeng2022lion} with DDIM sampling. Our experiments were conducted on an NVIDIA A100 GPU with the Flash Attention layer enabled \cite{dao2022flashattention}.

\noindent {\bf CFG scale factor }  Table \ref{tab:cfg} presents the ablation study on the CFG scaling factor $s_{\rm g}$ for 3D human mesh recovery from an image. For Human 3.6M, a value up to around 1 leads to improvements. On the other hand, we observed that increasing the CFG factor negatively impacts mesh reconstruction on the 3DPW dataset. This discrepancy is likely due to the difference in the accuracy of 3D pose regressors and the frequency of occlusions associated with each dataset. In general, the 3D joint estimation results on Human3.6M are more accurate and reliable than those on 3DPW.  This is because the Human3.6M dataset is captured in a controlled experimental environment and is included in the training, whereas 3DPW is an in-the-wild dataset and not used as the training data. DiffSurf provides a method to consider the accuracy and reliability of the 3D joint prediction by balancing between diffusion model priors and 3D body joint conditions through the CFG scaling factor.





\section{Conclusion}

We presented DiffSurf, a denoising diffusion transformer model for generating 3D surfaces in diverse body shapes and poses. DiffSurf can generate 3D surfaces for a wide range of object types and solve various downstream 3D processing tasks. In future work, we aim to extend DiffSurf towards the generation of expressive human body meshes with fine-grained details, such as facial expressions and finger poses. It would also be intriguing to design a foundational 3D generative model by increasing the capacity of DiffSurf and training it on a larger-scale 3D data.

\section*{Acknowledgements} 

This work was supported by JSPS KAKENHI Grant Number 20K19836,  22H00545, 23H03426 and 23K28116 in Japan. This work was supported by NEDO JPNP20006 (New Energy and Industrial Technology Development Organization) in Japan.

    \bibliographystyle{splncs04}
    \bibliography{main}

\begin{thebibliography}{10}
\providecommand{\url}[1]{\texttt{#1}}
\providecommand{\urlprefix}{URL }
\providecommand{\doi}[1]{https://doi.org/#1}

\bibitem{alliegro2023polydiff}
Alliegro, A., Siddiqui, Y., Tommasi, T., Nießner, M.: Polydiff: Generating 3d polygonal meshes with diffusion models (2023)

\bibitem{andriluka14cvpr}
Andriluka, M., Pishchulin, L., Gehler, P., Bernt, S.: 2d human pose estimation: New benchmark and state of the art analysis. In: CVPR (2014)

\bibitem{9010824}
Aumentado-Armstrong, T., Tsogkas, S., Jepson, A., Dickinson, S.: Geometric disentanglement for generative latent shape models. In: ICCV. pp. 8180--8189 (2019)

\bibitem{bao2022all}
Bao, F., Nie, S., Xue, K., Cao, Y., Li, C., Su, H., Zhu, J.: All are worth words: A vit backbone for diffusion models. In: CVPR (2023)

\bibitem{bao2022one}
Bao, F., Nie, S., Xue, K., Li, C., Pu, S., Wang, Y., Yue, G., Cao, Y., Su, H., Zhu, J.: One transformer fits all distributions in multi-modal diffusion at scale  (2023)

\bibitem{bautista2022gaudi}
Bautista, M.A., Guo, P., Abnar, S., Talbott, W., Toshev, A., Chen, Z., Dinh, L., Zhai, S., Goh, H., Ulbricht, D., Dehghan, A., Susskind, J.: Gaudi: A neural architect for immersive 3d scene generation. arXiv  (2022)

\bibitem{biggs2020multibodies}
Biggs, B., Ehrhart, S., Joo, H., Graham, B., Vedaldi, A., Novotny, D.: {3D} multibodies: Fitting sets of plausible {3D} models to ambiguous image data. In: NeurIPS (2020)

\bibitem{bogo2016keep}
Bogo, F., Kanazawa, A., Lassner, C., Gehler, P., Romero, J., Black, M.J.: Keep it smpl: Automatic estimation of 3d human pose and shape from a single image. In: ECCV. pp. 561--578. Springer (2016)

\bibitem{dfaust2017}
Bogo, F., Romero, J., Pons-Moll, G., Black, M.J.: Dynamic {FAUST}: {R}egistering human bodies in motion. In: CVPR (Jul 2017)

\bibitem{chang2015shapenet}
Chang, A.X., Funkhouser, T., Guibas, L., Hanrahan, P., Huang, Q., Li, Z., Savarese, S., Savva, M., Song, S., Su, H., Xiao, J., Yi, L., Yu, F.: Shapenet: An information-rich 3d model repository (2015)

\bibitem{iepgan2021}
Chen, H., Tang, H., Shi, H., Peng, W., Sebe, N., Zhao, G.: Intrinsic-extrinsic preserved gans for unsupervised 3d pose transfer. In: ICCV. pp. 8610--8619 (2021)

\bibitem{DBLP:journals/corr/abs-1903-10384}
Cheng, S., Bronstein, M.M., Zhou, Y., Kotsia, I., Pantic, M., Zafeiriou, S.: Meshgan: Non-linear 3d morphable models of faces. CoRR  \textbf{abs/1903.10384} (2019)

\bibitem{cheng2023sdfusion}
Cheng, Y.C., Lee, H.Y., Tulyakov, S., Schwing, A.G., Gui, L.Y.: {SDFusion}: Multimodal 3d shape completion, reconstruction, and generation. In: CVPR. pp. 4456--4465 (2023)

\bibitem{cho2023generative}
Cho, H., Kim, J.: Generative approach for probabilistic human mesh recovery using diffusion models (2023)

\bibitem{cho_arxiv.2207.13820}
Cho, J., Youwang, K., Oh, T.H.: Cross-attention of disentangled modalities for 3d human mesh recovery with transformers. In: ECCV (2022)

\bibitem{Choi_2020_ECCV_Pose2Mesh}
Choi, H., Moon, G., Lee, K.M.: Pose2mesh: Graph convolutional network for 3d human pose and mesh recovery from a 2d human pose. In: ECCV (2020)

\bibitem{Freihand2019}
Christian, Z., Duygu, C., Jimei, Y., Russel, B., Argus, M., Brox, T.: Freihand: A dataset for markerless capture of hand pose and shape from single rgb images. In: ICCV (2019)

\bibitem{LIMP2020}
Cosmo, L., Norelli, A., Halimi, O., Kimmel, R., Rodol\`{a}, E.: Limp: Learning latent shape representations with metric preservation priors. In: ECCV. p. 19–35 (2020)

\bibitem{dao2022flashattention}
Dao, T., Fu, D.Y., Ermon, S., Rudra, A., R{\'e}, C.: Flash{A}ttention: Fast and memory-efficient exact attention with {IO}-awareness. In: NeurIPS (2022)

\bibitem{9879900}
Davydov, A., Remizova, A., Constantin, V., Honari, S., Salzmann, M., Fua, P.: Adversarial parametric pose prior. In: CVPR. pp. 10987--10995 (jun 2022)

\bibitem{desbrun99}
Desbrun, M., Meyer, M., Schr\"{o}der, P., Barr, A.H.: Implicit fairing of irregular meshes using diffusion and curvature flow. In: Proceedings of the 26th Annual Conference on Computer Graphics and Interactive Techniques. p. 317–324. SIGGRAPH '99 (1999)

\bibitem{Frohlich2021}
Fröhlich, S., Botsch, M.: Example-driven deformations based on discrete shells. Computer Graphics Forum  \textbf{30}(8),  2246--2257 (2011)

\bibitem{Gao2021}
Gao, L., Lai, Y.K., Yang, J., Zhang, L.X., Xia, S., Kobbelt, L.: Sparse data driven mesh deformation. IEEE Transactions on Visualization and Computer Graphics  \textbf{27}(3),  2085--2100 (2021)

\bibitem{hierachical2020}
Georgakis, G., Li, R., Karanam, S., Chen, T., Ko{\v{s}}eck{\'a}, J., Wu, Z.: Hierarchical kinematic human mesh recovery. In: Vedaldi, A., Bischof, H., Brox, T., Frahm, J.M. (eds.) ECCV. pp. 768--784. Springer International Publishing (2020)

\bibitem{gong2023diffpose}
Gong, J., Foo, L.G., Fan, Z., Ke, Q., Rahmani, H., Liu, J.: Diffpose: Toward more reliable 3d pose estimation. In: CVPR (June 2023)

\bibitem{groueix2018b}
Groueix, T., Fisher, M., Kim, V.G., Russell, B., Aubry, M.: 3d-coded : 3d correspondences by deep deformation. In: ECCV (2018)

\bibitem{Ho2022ClassifierFreeDG}
Ho, J.: Classifier-free diffusion guidance. ArXiv  \textbf{abs/2207.12598} (2022)

\bibitem{ho2020denoising}
Ho, J., Jain, A., Abbeel, P.: Denoising diffusion probabilistic models. arXiv preprint arxiv:2006.11239  (2020)

\bibitem{h36m_pami}
Ionescu, C., Papava, D., Olaru, V., Sminchisescu, C.: Human3.6m: Large scale datasets and predictive methods for 3d human sensing in natural environments. IEEE TPAMI  \textbf{36}(7),  1325--1339 (2014)

\bibitem{Jiang2020HumanBody}
Jiang, B., Zhang, J., Cai, J., Zheng, J.: Disentangled human body embedding based on deep hierarchical neural network  (2020)

\bibitem{Johnson2011}
Johnson, S., Everingham, M.: Learning effective human pose estimation from inaccurate annotation. In: CVPR. pp. 1465--1472 (2011)

\bibitem{joo2020eft}
Joo, H., Neverova, N., Vedaldi, A.: Exemplar fine-tuning for 3d human pose fitting towards in-the-wild 3d human pose estimation. In: 3DV (2020)

\bibitem{jun2023shape}
Jun, H., Nichol, A.: Shap-e: Generating conditional 3d implicit functions (2023)

\bibitem{hmrKanazawa17}
Kanazawa, A., Black, M.J., Jacobs, D.W., Malik, J.: End-to-end recovery of human shape and pose. In: CVPR (2018)

\bibitem{Khirodkar_2022_CVPR}
Khirodkar, R., Tripathi, S., Kitani, K.: Occluded human mesh recovery. In: CVPR. pp. 1715--1725 (June 2022)

\bibitem{Kocabas_PARE_2021}
Kocabas, M., Huang, C.H.P., Hilliges, O., Black, M.J.: {PARE}: Part attention regressor for {3D} human body estimation. In: ICCV. pp. 11127--11137 (2021)

\bibitem{kolotouros2019spin}
Kolotouros, N., Pavlakos, G., Black, M.J., Daniilidis, K.: Learning to reconstruct 3d human pose and shape via model-fitting in the loop. In: ICCV (2019)

\bibitem{kolotouros2019cmr}
Kolotouros, N., Pavlakos, G., Daniilidis, K.: Convolutional mesh regression for single-image human shape reconstruction. In: CVPR (2019)

\bibitem{kolotouros2021prohmr}
Kolotouros, N., Pavlakos, G., Jayaraman, D., Daniilidis, K.: Probabilistic modeling for human mesh recovery. In: ICCV (2021)

\bibitem{li2023diffhand}
Li, L., Zhuo, L., Zhang, B., Bo, L., Chen, C.: Diffhand: End-to-end hand mesh reconstruction via diffusion models (2023)

\bibitem{lin2023magic3d}
Lin, C.H., Gao, J., Tang, L., Takikawa, T., Zeng, X., Huang, X., Kreis, K., Fidler, S., Liu, M.Y., Lin, T.Y.: Magic3d: High-resolution text-to-3d content creation. In: CVPR (2023)

\bibitem{dat2023}
Lin, G.F., Jia, G., Hossein, R., Jun, L.: Distribution-aligned diffusion for human mesh recovery. In: ICCV (2023)

\bibitem{lin2021end-to-end}
Lin, K., Wang, L., Liu, Z.: End-to-end human pose and mesh reconstruction with transformers. In: CVPR (2021)

\bibitem{lin2021-mesh-graphormer}
Lin, K., Wang, L., Liu, Z.: Mesh graphormer. In: ICCV (2021)

\bibitem{LinMBHPRDZ14}
Lin, T., Maire, M., Belongie, S.J., Bourdev, L.D., Girshick, R.B., Hays, J., Perona, P., Ramanan, D., Doll{\'{a}}r, P., Zitnick, C.L.: Microsoft {COCO:} common objects in context. CoRR  \textbf{abs/1405.0312} (2014)

\bibitem{10161247}
Liu, Y., Yang, J., Gu, X., Guo, Y., Yang, G.Z.: Egohmr: Egocentric human mesh recovery via hierarchical latent diffusion model. In: 2023 IEEE International Conference on Robotics and Automation (ICRA). pp. 9807--9813 (2023)

\bibitem{Liu2023MeshDiffusion}
Liu, Z., Feng, Y., Black, M.J., Nowrouzezahrai, D., Paull, L., Liu, W.: Meshdiffusion: Score-based generative 3d mesh modeling. In: International Conference on Learning Representations (2023)

\bibitem{SMPL:2015}
Loper, M., Mahmood, N., Romero, J., Pons-Moll, G., Black, M.J.: {SMPL}: A skinned multi-person linear model. ACM TOG  \textbf{34}(6),  248:1--248:16 (2015)

\bibitem{luo2021diffusion}
Luo, S., Hu, W.: Diffusion probabilistic models for 3d point cloud generation. In: CVPR (June 2021)

\bibitem{lyu2022conditional}
Lyu, Z., Kong, Z., Xu, X., Pan, L., Lin, D.: A conditional point diffusion-refinement paradigm for 3d point cloud completion (2022)

\bibitem{slide2023}
Lyu, Z., Wang, J., An, Y., Zhang, Y., Lin, D., Dai, B.: Controllable mesh generation through sparse latent point diffusion models. CVPR (2023)

\bibitem{ma2020cape}
Ma, Q., Yang, J., Ranjan, A., Pujades, S., Pons-Moll, G., Tang, S., Black, M.J.: Learning to dress 3d people in generative clothing. In: CVPR (Jun 2020)

\bibitem{AMASS:2019}
Mahmood, N., Ghorbani, N., F.~Troje, N., Pons-Moll, G., Black, M.J.: Amass: Archive of motion capture as surface shapes. In: ICCV (2019)

\bibitem{vonMarcard2018}
von Marcard, T., Henschel, R., Black, M., Rosenhahn, B., Pons-Moll, G.: Recovering accurate 3d human pose in the wild using imus and a moving camera. In: ECCV (2018)

\bibitem{mono-3dhp2017}
Mehta, D., Rhodin, H., Casas, D., Fua, P., Sotnychenko, O., Xu, W., Theobalt, C.: Monocular 3d human pose estimation in the wild using improved cnn supervision. In: 3DV. IEEE (2017)

\bibitem{mo2023dit3d}
Mo, S., Xie, E., Chu, R., Hong, L., Nießner, M., Li, Z.: Dit-3d: Exploring plain diffusion transformers for 3d shape generation. arXiv preprint arXiv: 2307.01831  (2023)

\bibitem{nichol2022pointe}
Nichol, A., Jun, H., Dhariwal, P., Mishkin, P., Chen, M.: Point-e: A system for generating 3d point clouds from complex prompts (2022)

\bibitem{SMPL-X:2019}
Pavlakos, G., Choutas, V., Ghorbani, N., Bolkart, T., Osman, A.A.A., Tzionas, D., Black, M.J.: Expressive body capture: 3d hands, face, and body from a single image. In: CVPR (2019)

\bibitem{Peebles2022}
Peebles, W., Radosavovic, I., Brooks, T., Efros, A., Malik, J.: Learning to learn with generative models of neural network checkpoints. arXiv preprint arXiv:2209.12892  (2022)

\bibitem{Peebles2022DiT}
Peebles, W., Xie, S.: Scalable diffusion models with transformers. arXiv preprint arXiv:2212.09748  (2022)

\bibitem{Peng2021SAP}
Peng, S., Jiang, C.M., Liao, Y., Niemeyer, M., Pollefeys, M., Geiger, A.: Shape as points: A differentiable poisson solver. In: NeurIPS (2021)

\bibitem{lucidrain}
Phil, W.: denoising-diffusion-pytorch. \url{https://github.com/lucidrains/denoising-diffusion-pytorch} (2023)

\bibitem{poole2022dreamfusion}
Poole, B., Jain, A., Barron, J.T., Mildenhall, B.: Dreamfusion: Text-to-3d using 2d diffusion. arXiv  (2022)

\bibitem{COMA:ECCV18}
Ranjan, A., Bolkart, T., Sanyal, S., Black, M.J.: Generating {3D} faces using convolutional mesh autoencoders. In: ECCV. pp. 725--741 (2018)

\bibitem{Rombach_2022_CVPR}
Rombach, R., Blattmann, A., Lorenz, D., Esser, P., Ommer, B.: High-resolution image synthesis with latent diffusion models. In: CVPR. pp. 10684--10695 (June 2022)

\bibitem{MANO:SIGGRAPHASIA:2017}
Romero, J., Tzionas, D., Black, M.J.: Embodied hands: Modeling and capturing hands and bodies together. ACM TOG  \textbf{36}(6) (Nov 2017)

\bibitem{BARC:2022}
Rueegg, N., Zuffi, S., Schindler, K., Black, M.J.: Barc: Learning to regress 3d dog shape from images by exploiting breed information. In: CVPR (2022)

\bibitem{salimans2022progressive}
Salimans, T., Ho, J.: Progressive distillation for fast sampling of diffusion models (2022)

\bibitem{shan2023diffusion}
Shan, W., Liu, Z., Zhang, X., Wang, Z., Han, K., Wang, S., Ma, S., Gao, W.: Diffusion-based 3d human pose estimation with multi-hypothesis aggregation. arXiv preprint arXiv:2303.11579  (2023)

\bibitem{shen2021dmtet}
Shen, T., Gao, J., Yin, K., Liu, M.Y., Fidler, S.: Deep marching tetrahedra: a hybrid representation for high-resolution 3d shape synthesis. In: Advances in Neural Information Processing Systems (NeurIPS) (2021)

\bibitem{shim2023diffusion}
Shim, J., Kang, C., Joo, K.: Diffusion-based signed distance fields for 3d shape generation. In: CVPR. pp. 20887--20897 (2023)

\bibitem{slerp}
Shoemake, K.: Animating rotation with quaternion curves. In: Proceedings of the 12th Annual Conference on Computer Graphics and Interactive Techniques. p. 245–254. SIGGRAPH '85, Association for Computing Machinery, New York, NY, USA (1985)

\bibitem{song2020denoising}
Song, J., Meng, C., Ermon, S.: Denoising diffusion implicit models. arXiv:2010.02502  (October 2020)

\bibitem{lgd2020}
Song, J., Chen, X., Hilliges, O.: Human body model fitting by learned gradient descent. In: Vedaldi, A., Bischof, H., Brox, T., Frahm, J.M. (eds.) ECCV. pp. 744--760 (2020)

\bibitem{SumnerZGP05}
Sumner, R.W., Zwicker, M., Gotsman, C., Popovic, J.: Mesh-based inverse kinematics. ACM TOG  \textbf{24}(3),  488--495 (2005)

\bibitem{SemanticHuman}
Sun, X., Feng, Q., Li, X., Zhang, J., Lai, Y.K., Yang, J., Li, K.: Learning semantic-aware disentangled representation for flexible 3d human body editing. In: CVPR (2023)

\bibitem{8578710}
Tan, Q., Gao, L., Lai, Y.K., Xia, S.: Variational autoencoders for deforming 3d mesh models. In: CVPR. pp. 5841--5850 (2018)

\bibitem{tian2022hmrsurvey}
Tian, Y., Zhang, H., Liu, Y., Wang, L.: Recovering 3d human mesh from monocular images: A survey. arXiv preprint arXiv:2203.01923  (2022)

\bibitem{tiwari22posendf}
Tiwari, G., Antic, D., Lenssen, J.E., Sarafianos, N., Tung, T., Pons-Moll, G.: Pose-ndf: Modeling human pose manifolds with neural distance fields. In: ECCV (October 2022)

\bibitem{varol17_surreal}
Varol, G., Romero, J., Martin, X., Mahmood, N., Black, M.J., Laptev, I., Schmid, C.: Learning from synthetic humans. In: CVPR (2017)

\bibitem{wang2023prolificdreamer}
Wang, Z., Lu, C., Wang, Y., Bao, F., Li, C., Su, H., Zhu, J.: Prolificdreamer: High-fidelity and diverse text-to-3d generation with variational score distillation. arXiv preprint arXiv:2305.16213  (2023)

\bibitem{xu2020ghum}
Xu, H., Bazavan, E.G., Zanfir, A., Freeman, W.T., Sukthankar, R., Sminchisescu, C.: Ghum \& ghuml: Generative 3d human shape and articulated pose models. In: CVPR. pp. 6184--6193 (2020)

\bibitem{xu2023animal3d}
Xu, J., Zhang, Y., Peng, J., Ma, W., Jesslen, A., Ji, P., Hu, Q., Zhang, J., Liu, Q., Wang, J., et~al.: Animal3d: A comprehensive dataset of 3d animal pose and shape. arXiv preprint arXiv:2308.11737  (2023)

\bibitem{Xu_2023_ICCV}
Xu, X., Wang, Z., Zhang, G., Wang, K., Shi, H.: Versatile diffusion: Text, images and variations all in one diffusion model. In: ICCV. pp. 7754--7765 (October 2023)

\bibitem{pointflow}
Yang, G., Huang, X., Hao, Z., Liu, M.Y., Belongie, S., Hariharan, B.: Pointflow: 3d point cloud generation with continuous normalizing flows. arXiv  (2019)

\bibitem{yoshiyasu2023-deformer}
Yoshiyasu, Y.: Deformable mesh transformer for 3d human mesh recovery. In: CVPR. pp. 17006--17015 (2023)

\bibitem{10096870}
You, Y., Liu, H., Li, X., Li, W., Wang, T., Ding, R.: Gator: Graph-aware transformer with motion-disentangled regression for human mesh recovery from a 2d pose. In: ICASSP 2023 - 2023 IEEE International Conference on Acoustics, Speech and Signal Processing (ICASSP). pp.~1--5 (2023)

\bibitem{yu2023surf}
Yu, Z., Dou, Z., Long, X., Lin, C., Li, Z., Liu, Y., Müller, N., Komura, T., Habermann, M., Theobalt, C., et~al.: Surf-d: High-quality surface generation for arbitrary topologies using diffusion models. arXiv preprint arXiv:2311.17050  (2023)

\bibitem{yuan2020mesh}
Yuan, Y.J., Lai, Y.K., Yang, J., Duan, Q., Fu, H., Gao, L.: Mesh variational autoencoders with edge contraction pooling. In: CVPRW. pp. 274--275 (2020)

\bibitem{zanfir2020weakly}
Zanfir, A., Bazavan, E.G., Xu, H., Freeman, W.T., Sukthankar, R., Sminchisescu, C.: Weakly supervised 3d human pose and shape reconstruction with normalizing flows. In: ECCV. pp. 465--481 (2020)

\bibitem{zeng2022lion}
Zeng, X., Vahdat, A., Williams, F., Gojcic, Z., Litany, O., Fidler, S., Kreis, K.: Lion: Latent point diffusion models for 3d shape generation. In: Advances in Neural Information Processing Systems (NeurIPS) (2022)

\bibitem{pymaf2021}
Zhang, H., Tian, Y., Zhou, X., Ouyang, W., Liu, Y., Wang, L., Sun, Z.: Pymaf: 3d human pose and shape regression with pyramidal mesh alignment feedback loop. In: ICCV (2021)

\bibitem{zhou20unsupervised}
Zhou, K., Bhatnagar, B.L., Pons-Moll, G.: Unsupervised shape and pose disentanglement for 3d meshes. In: ECCV (August 2020)

\bibitem{Zuffi:CVPR:2017}
Zuffi, S., Kanazawa, A., Jacobs, D., Black, M.J.: {3D} menagerie: Modeling the {3D} shape and pose of animals. In: CVPR (Jul 2017)

\end{thebibliography}



\end{document}